\documentclass[letterpaper, 10 pt, conference, nofonttune]{ieeeconf}  

\IEEEoverridecommandlockouts                              

\usepackage{graphics} 
\usepackage{epsfig} 
\usepackage{mathptmx} 
\usepackage{times} 
\usepackage{amsmath} 
\usepackage{amssymb}  
\usepackage{bm}
\usepackage{caption}
\usepackage{cite} 
\usepackage{algorithm}
\usepackage{booktabs}
\usepackage{algpseudocode}
\DeclareSymbolFont{cmsymbols}{OMS}{cmsy}{m}{n}
\SetSymbolFont{cmsymbols}{bold}{OMS}{cmsy}{b}{n}
\DeclareSymbolFontAlphabet{\mathcal}{cmsymbols}
\usepackage{multirow}   

\title{\LARGE \bf
Real-time Whole-Body Motion Planning for Mobile Manipulators Carrying Arbitrarily Shaped Payloads via Kinematically-Coupled SVSDF
}

\author{Yisheng Li*$^{1\,}$, Longji Yin*$^{1\,}$, Tingrui Zhang$^2$, Ruize Xue$^1$, Haoda Zhu$^1$, Nan Chen$^1$, Siqi Liang$^1$,\\
Yuxi Liu$^1$, Fu Zhang$^1$%
\thanks{\textbf{${*}$ Equal contribution.}}
\thanks{$^1$Y. Li, L. Yin, R. Xue, H. Zhu, N. Chen, S. Liang, Y. Liu and F. Zhang are with the Department of Mechanical Engineering, University of Hong Kong.}%
\thanks{$^2$T. Zhang, is with Institute for Interdisciplinary Information Sciences, Tsinghua University.}
\thanks{\raggedright Email:{\tt\footnotesize \{yli385,ljyin,u3603390,haodazhumars,\newline cnchen,liangsiqi,liuyuxi\}@connect.hku.hk}, {\tt\footnotesize m15351201779@163.com}, {\tt\footnotesize fuzhang@hku.hk}.}
\thanks{Corresponding Author: Fu Zhang.}
}

\begin{document}

\graphicspath{{Pictures/}}  

\maketitle

\thispagestyle{empty}
\pagestyle{empty}

\begin{abstract}

Mobile manipulators are increasingly tasked with transporting large, non-convex payloads through cluttered environments, yet existing planners either oversimplify the payload geometry or fail to handle the kinematic coupling between manipulator links, leading to lost feasible space or stalled optimization. This letter presents a real-time whole-body motion planning framework for mobile manipulators carrying arbitrarily shaped payloads. The front-end employs a chain-decomposed kernel-based collision check that preserves the true geometry of the robot and payload, with compact storage and fast bit-level queries. A mid-end preprocessing stage converts the front-end path into a continuous trajectory enforcing smoothness and feasibility, and executes it directly when collision-free to bypass the costly back-end. When refinement is required, the back-end performs trajectory optimization built on a Kinematically-Coupled SVSDF (KC-SVSDF), which propagates collision-avoidance gradients along the kinematic chain to produce coherent whole-body escape directions. Ablation studies, comparative benchmarks against state-of-the-art baselines, and real-world experiments on a differential-drive mobile manipulator demonstrate that the proposed framework reliably transports large, non-convex payloads through tight passages and cluttered environments.

\end{abstract}


\section{INTRODUCTION}

Recent advances in joint motor technology have substantially increased the payload capacity of mobile manipulators. This expanded capability opens up new application scenarios, such as relocating pianos or sofas in domestic settings and handling die-cast aluminum car doors on factory floors. In such tasks, however, the grasped payload is typically large and non-convex, and conventional planners that treat it as a small object and ignore its actual geometry render these applications unsafe. This letter addresses the problem of real-time whole-body motion planning for a mobile manipulator transporting a large, non-convex payload through cluttered environments.

\begin{figure}[t]
    \centering
    \includegraphics[width=0.48\textwidth]{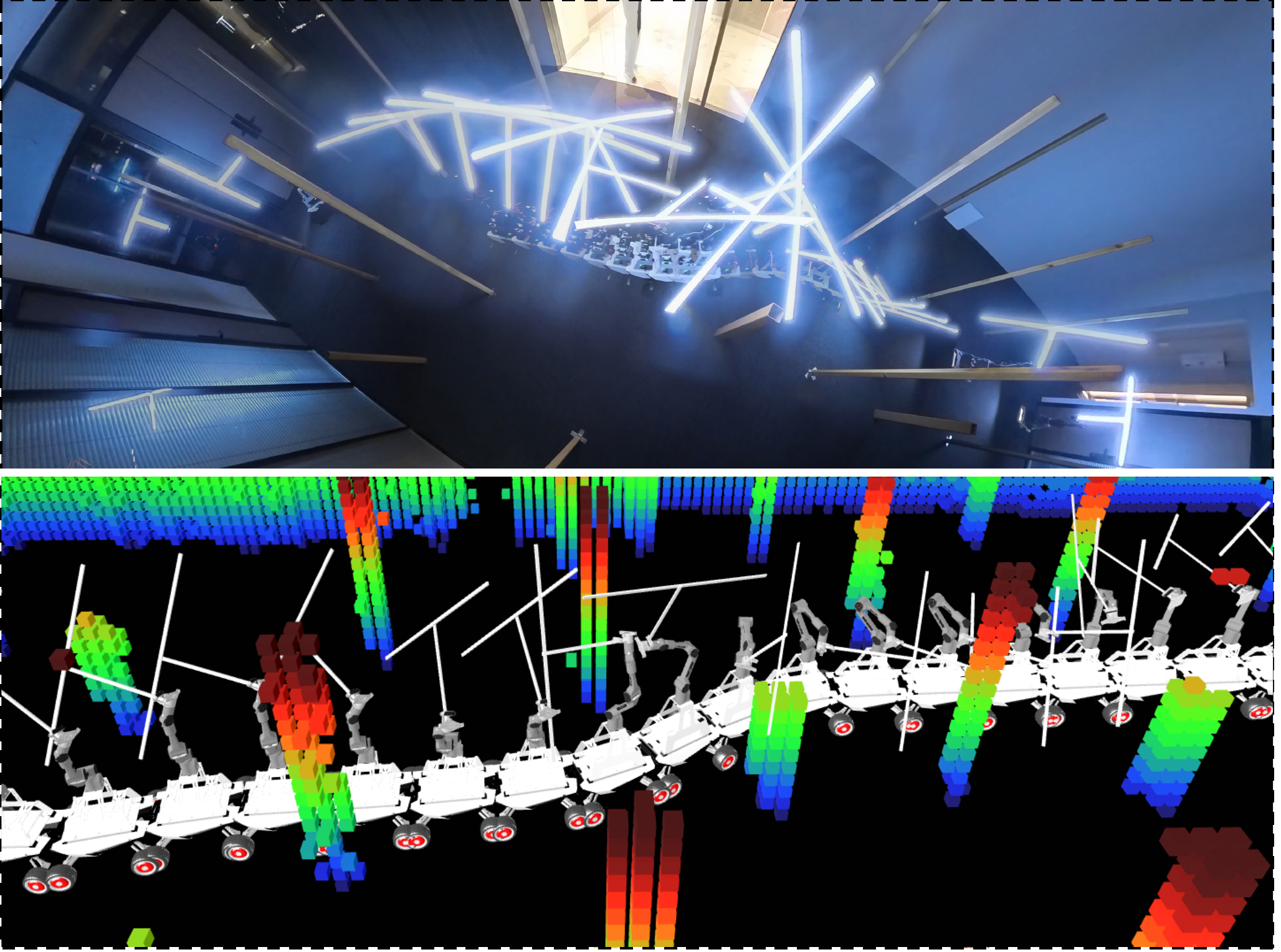}
    \captionsetup{font={footnotesize}}
    \caption{Real-world experiment1: the mobile manipulator traverses a cluttered environment of randomly placed wooden columns while grasping a large T-shaped carbon-fiber payload, demonstrating safe whole-body navigation through tight gaps with an arbitrarily shaped object.}
    \label{fig:cover_page}
    \vspace{-20pt}
\end{figure}

Generating a real-time, feasible, and high-quality trajectory under these conditions remains a significant challenge. Existing optimization-based mobile manipulator planners~\cite{remani, topay} approximate both the robot body and the grasped payload by a set of bounding spheres and rely on ESDF queries for collision evaluation. This approximation is acceptable for small payloads such as a water bottle, but breaks down for large, non-convex objects: a bounding sphere cover sacrifices feasible space and often fails to find any valid path. To support arbitrarily shaped robots, Zhang~\textit{et~al.}~\cite{SVSDF} proposed the Swept-Volume Signed Distance Field (SVSDF), but the formulation is derived for a single rigid body. Applying it independently to each link of a serial manipulator causes problems at both ends of the pipeline: in the front-end, enumerating a kernel for every joint-sample combination yields an intractable storage cost; in the back-end, the per-link escape gradients ignore the rigid coupling along the kinematic chain, so the direction that frees one link can drive an upstream link into a new collision and stall the optimizer.

To overcome these limitations, we propose a real-time whole-body motion planning framework for mobile manipulators carrying arbitrarily shaped payloads. The framework consists of three modules: (i) a front-end path search using a chain-decomposed kernel-based collision check that preserves the true robot and payload geometry, with compact storage and fast bit-level queries; (ii) a mid-end preprocessing stage that, provisionally trusting the safety of the front-end path, converts it into a continuous trajectory enforcing only smoothness and feasibility, and executes it directly when collision-free; and (iii) a back-end trajectory optimization built on a Kinematically-Coupled SVSDF (KC-SVSDF), which propagates collision-avoidance gradients along the kinematic chain to produce coherent whole-body escape directions. The framework is validated through ablation studies, comparative benchmarks against state-of-the-art baselines, and real-world experiments on a differential-drive mobile manipulator.
The main contributions of this letter are:
\begin{itemize}
\item We proposed a chain-decomposed kernel-based collision-checking scheme that preserves the true geometry of the mobile manipulator and its grasped payload, with storage scaling linearly in the per-joint sample counts rather than multiplicatively.
\item We develop a trajectory generation framework centered on hierarchical optimization that enables real-time, continuous collision safety for wheeled mobile manipulator while grasping arbitrarily shaped payloads.
\item We proposed a Kinematically-Coupled SVSDF (KC-SVSDF) method, which coupled multi-link systems by enforcing consistency of per-link escape directions along the kinematic chain.
\item We demonstrate the effectiveness of the proposed  method in simulation and real-world experiments. The code will be publicly released.
\end{itemize}


\section{RELATED WORKS}
Path planning for mobile manipulators is particularly challenging when the system transports a large, non-convex payload through cluttered environments under real-time re-planning requirements. The high dimensionality of the whole-body configuration space, the heterogeneous kinematics of the mobile base and the manipulator, and the non-convex geometry of the grasped object jointly inflate the search space and complicate collision checking. Existing approaches can be broadly grouped into three categories.

\textit{Decoupled methods}~\cite{decouple2-HAMP, decouple3} plan the base and the arm sequentially or hierarchically, so that each sub-problem lives in a low-dimensional space. The strict ordering between the two stages, however, yields sub-optimal plans and often fails in narrow passages where base and arm motions must be tightly coordinated. \textit{Whole-body samplers}~\cite{wholebody1, wholebody2, wholebody3, wholebody4} instead plan directly in the joint configuration space of the base and the arm, recovering the lost coupling at the cost of dimensionality that makes them too slow for online re-planning. Kang~\textit{et~al.}~\cite{hybrid-coupled} bridge the two paradigms with a hybrid scheme that decouples base and manipulator sampling in open space and switches to joint sampling only in highly cluttered regions; the method nevertheless inherits the slow convergence of random sampling. Furthermore, their approach simplifies the true geometry of the robot body and payloads, which prevents finding feasible solutions in cluttered environments.

\textit{Optimization methods}~\cite{remani, chomp, topay} first computes a discrete whole-body initial path and then optimize it with a whole-body trajectory optimizer to produce a smooth, continues, feasible trajectory. These methods approximate the mobile manipulator by a set of bounding spheres queried against a signed distance field, and absorb the grasped payload into the same sphere-based surrogate during both the front-end search and the back-end optimization. When the payload is large and non-convex, this approximation inflates the effective collision geometry: feasible paths become hard to find in the front end, and the back-end optimizer is prone to local minima, ultimately precluding a smooth, kinematically feasible, and continuously collision-free trajectory. Zhang~\textit{et~al.}~\cite{SVSDF,svsdf_iros} address the geometry issue with a swept-volume SDF that supports arbitrarily shaped robot. The formulation, however, is derived for a single rigid body: applied independently to each link of a serial manipulator, it produces per-link escape gradients that ignore the rigid-body coupling along the kinematic chain, so the direction that frees one link can drive an upstream link into a new collision, stalling the optimizer.

\section{PRELIMINARY}

\subsection{Kinematic Model} \label{sec:kinematics}
In this letter, we consider a mobile manipulator system consisting of a differential-drive mobile platform and a $J$-DOF manipulator mounted on top of the platform. The state of the mobile manipulator can be represented as $x = [x_b^\top, x_m^\top]^\top$, where $x_b = [p_x, p_y, \phi]^\top \in \mathbb{R}^3$ is the position and orientation of the mobile base and $x_m = [q_1, q_2, \ldots, q_J]^\top \in \mathbb{R}^J$ is the joint angle vector of the manipulator, whose end-effector rigidly grasps an object to be transported. The control input consists of wheel and joint angular velocities, $\mathbf{u} = [\omega_l, \omega_r, \dot{x}_m^\top]^\top$, where $\omega_l$ and $\omega_r$ are the left and right wheel angular velocities, and $\dot{x}_m$ is the joint velocity vector.


Combining the base flat output with the manipulator joint configuration $x_m$, we define the unified trajectory parameterization variable $z := [\sigma_b^\top, x_m^\top]^\top = [p_x, p_y, q_1, \ldots, q_J]^\top \in \mathbb{R}^{2+J}$, where the orientation is from $\phi = \operatorname{atan2}(\eta\dot{p}_y, \eta\dot{p}_x)$. The kinematic model of the base can be expressed as:
\vspace{-5pt}
\begin{subequations}\label{eq:kinematics}
\begin{align}
&\omega_{l(r)} = \frac{1}{2r_w} \left( 2\eta v_b \underset{(+)}{-} d_w \omega_b \right), \label{eq:1a} \\
&\dot{\omega}_{l(r)} = \alpha_{l(r)} = \frac{1}{2r_w} \left( 2\eta a_b \underset{(+)}{-} d_w \alpha_b \right), \label{eq:1b}
\end{align}
\end{subequations}
where $r_w$ is the radius of the driving wheels, $d_w$ is the wheelbase, and $\eta = \pm 1$ is a direction indicator ($+1$ for forward motion, $-1$ for backward motion). The quantities $v_b$, $a_b$, $\omega_b$, and $\alpha_b$ denote the linear velocity, linear acceleration, angular velocity, and angular acceleration of the mobile base, respectively~\cite{mengke,mengke_universe}.

\subsection{Trajectory parameterization}
We use MINCO~\cite{wang2022minco} for parameterization. The trajectory is partitioned into $N$ segments, where the $i$-th segment carries a direction indicator $\eta_i \in \{-1, +1\}$ and is divided into $M_i$ polynomial pieces of degree $2s-1$. The $j$-th piece of segment $i$ is defined as
\begin{equation}\label{eq:traj_piece}
z_{i,j}(t) = \mathbf{c}_{i,j}^\top \beta(t - \bar{T}_{i,j-1}), \quad t \in [\bar{T}_{i,j-1}, \bar{T}_{i,j}],
\end{equation}
where $\mathbf{c}_{i,j} \in \mathbb{R}^{2s \times (2+J)}$ are the polynomial coefficients, $\beta(t) = [1, t, \ldots, t^{2s-1}]^\top$ is the time basis, and $\bar{T}_{i,j} = \sum_{k=1}^{j} T_{i,k}$ is the cumulative time from the duration allocation $\mathbf{T}_i = [T_{i,1}, \ldots, T_{i,M_i}]^\top$.

\subsection{System Overview}
\begin{figure}[h]
    \centering
    \includegraphics[width=0.48\textwidth]{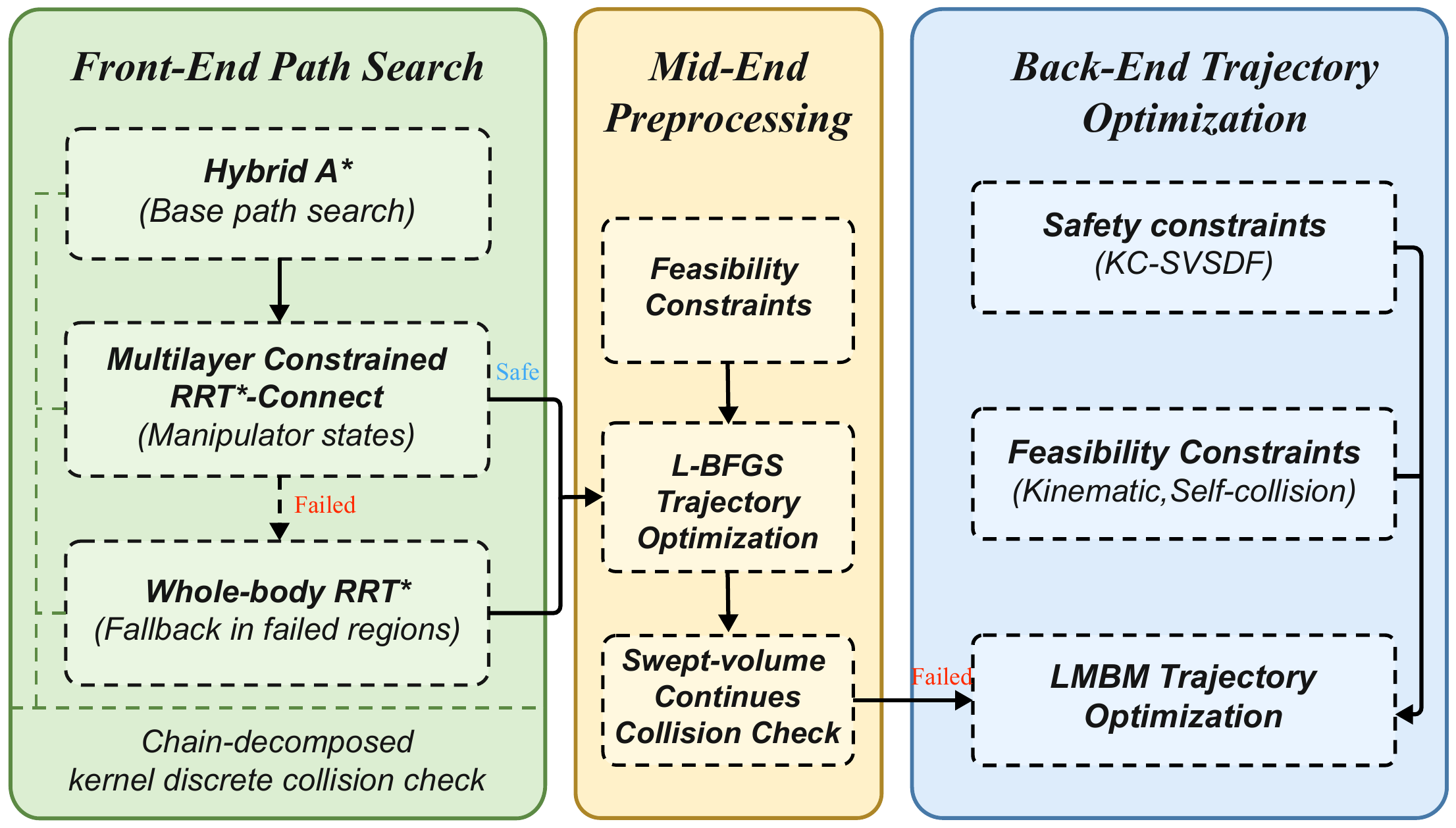}
    \captionsetup{font={footnotesize}}
    \caption{Overview of the proposed planning framework.}
    \label{fig:system_overview}
    \vspace{-10pt}
\end{figure}

In this letter, we propose a trajectory generation framework for payload-carrying mobile manipulators.
As illustrated in Fig.~\ref{fig:system_overview}, the framework consists of three hierarchical modules. \textbf{1) Front-End} (Sec.~\ref{sec:front-end}): To accommodate grasped objects of arbitrary shape, we employ a kernel-based fast collision-checking algorithm that integrates Hybrid A*~\cite{hybrid_a_star} for the mobile base with a Multilayer Constrained RRT*-Connect~\cite{remani} for the manipulator. When this decoupled search fails, a whole-body RRT* planner is invoked as a fallback to recover a feasible path. \textbf{2) Mid-End} (Sec.~\ref{sec:mid-end}): The discrete whole-body waypoints produced by the front-end are converted into a continuous trajectory. The resulting trajectory is either directly executed if collision-free, or otherwise serves as an initial guess for the back-end. \textbf{3) Back-End} (Sec.~\ref{sec:back-end}): Using the Kinematically-Coupled SVSDF (KC-SVSDF) together with the initialization from the mid-end, we perform whole-body trajectory optimization that jointly enforces continuous collision avoidance for the arbitrarily shaped payload and the system's feasibility.

\section{Front-end Path Search} \label{sec:front-end}

\subsection{Path Search Pipeline}
Our front-end pipeline follows a decoupled search strategy: Hybrid A*~\cite{hybrid_a_star} generates a base path under nonholonomic constraints, and a Multilayer Constrained RRT*-Connect~\cite{remani} determines the manipulator configuration at each base waypoint, with a whole-body RRT* invoked as a fallback when this decoupled search fails. The output is a collision-free path partitioned into segments with direction indicators $\eta_i \in \{-1, +1\}$ for forward and reverse motions, which serves as the initial guess for the mid-end. While these search algorithms are well established, their efficiency and safety hinge on the underlying collision-checking routine, which is the focus of this section.

\subsection{Existing Collision Checking}
Many existing mobile manipulator planners~\cite{remani, topay, NMPC_IROS} check collisions in the front-end by querying an ESDF map at points sampled on the robot body. Once an object is grasped, it is rigidly attached to the end-effector and should logically be treated as part of the robot, yet most planners reduce it to a bounding sphere or a convex surrogate. This point-and-primitive approximation discards the object's true geometry; to compensate, planners inflate the surrogate conservatively, but the over-inflation frequently rejects feasible configurations, especially when the payload is large and non-convex.

A recent approach~\cite{SVSDF, li2025efficient} avoids this approximation by encoding the robot and the environment on a common voxel grid as a pair of bit-packed kernels: a \emph{robot kernel} $\mathcal{K}_{robot}$ that captures the true robot geometry, and a \emph{map kernel} $\mathcal{K}_{map}$ that mirrors the occupancy voxel map, with its boundary voxels dilated outward by half the maximum extent of $\mathcal{K}_{robot}$ on each side. This representation preserves arbitrary geometry, and a collision query reduces to a fast boolean convolution between the two kernels. The formulation, however, assumes a single rigid body. A serial manipulator changes shape with its joint configuration, so adapting the original form directly would require enumerating one kernel per configuration, and the number of stored kernels grows as $O(\prod_{j=0}^{J} N_j)$ in the per-joint sample counts $N_j$, quickly becoming intractable.
\begin{figure}[h]
    \centering
    \includegraphics[width=0.48\textwidth]{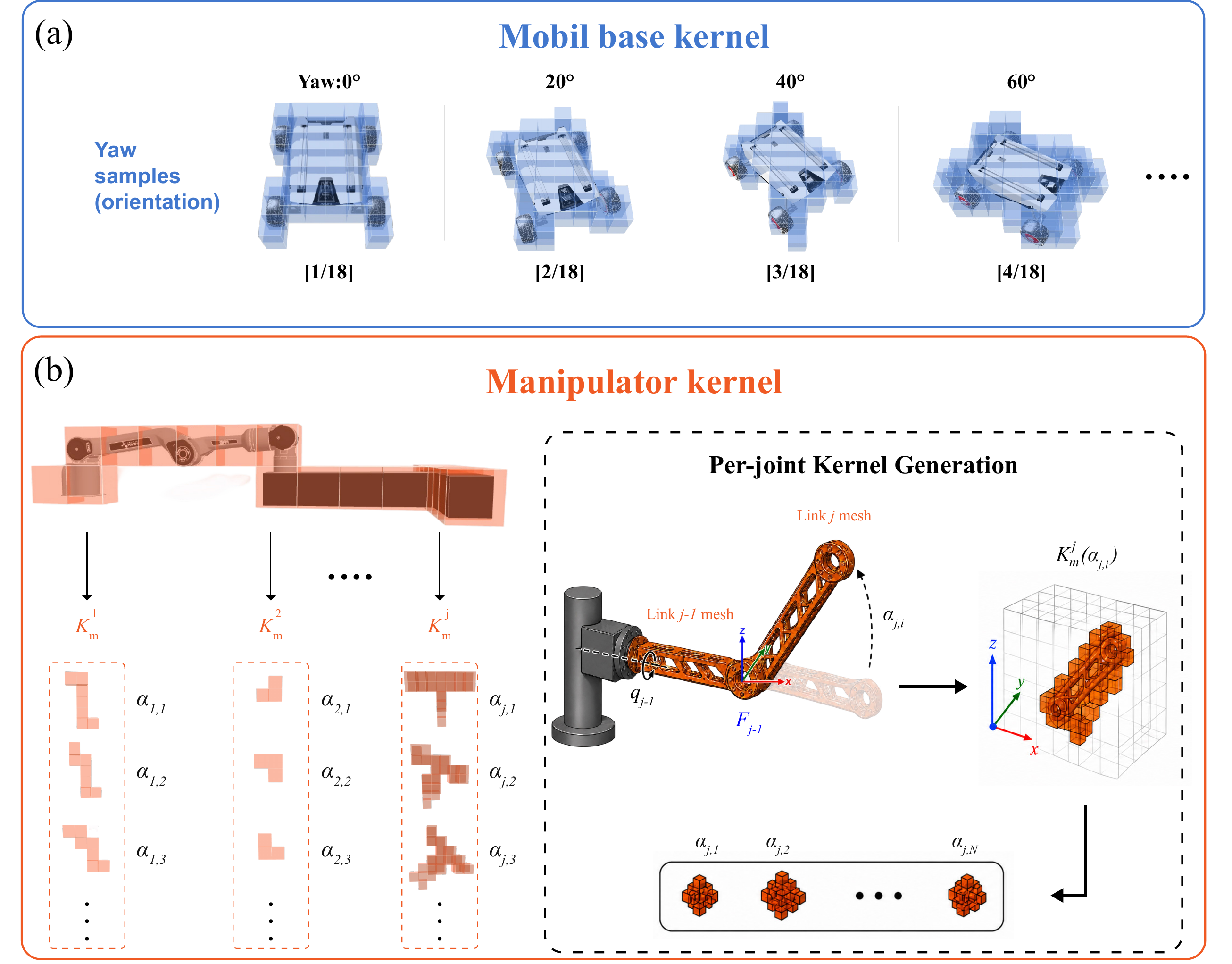}
    \captionsetup{font={footnotesize}}
    \caption{Chain-decomposed kernel generation. (a) The mobile base kernel is voxelized over discretized yaw samples in the world frame. (b) Each manipulator link kernel $\mathcal{K}_m^j$ is voxelized in its parent-link frame $F_{j-1}$ over discretized samples $\{\alpha_{j,1}, \ldots, \alpha_{j,N_j}\}$ of its local joint variable, with the grasped payload rigidly attached to the end-effector link.}
    \label{fig:kernelgeneration}
    \vspace{-10pt}
\end{figure}
\subsection{Chain-Decomposed Kernel} \label{sec:kernel-decomp}
To address this, we decompose the robot kernel into one kernel per link along the kinematic chain and resolve queries through forward kinematics, making the bit-packed representation tractable for a serial-link mobile manipulator. Concretely, we write $\mathcal{K}_{robot} = \{\mathcal{K}^0, \mathcal{K}^1_m, \ldots, \mathcal{K}^{J+1}_m\}$, treating the mobile base as $\mathcal{K}^0$ with its yaw angle $q_0$ as the local joint variable, the manipulator links as $\mathcal{K}^1_m$ through $J$ with local joint angles $q_1, \ldots, q_J$, and the grasped payload as link $J{+}1$ rigidly attached to the end-effector with no additional joint variable. As shown in Fig.~\ref{fig:kernelgeneration}, each $\mathcal{K}^j_m$ is voxelized in its parent-link frame $F_{j-1}$ over discretized samples $\{\alpha_{j,1}, \alpha_{j,2}, \ldots, \alpha_{j,N_j}\}$ of its local joint variable. At collision query time, given a base pose $x_b = [p_x, p_y, q_0]^\top$ and a manipulator configuration $x_m = [q_1, \dots, q_J]^\top$, we retrieve the precomputed kernel for each link of $\mathcal{K}_{robot}$ by selecting the sample $\alpha_j$ closest to the corresponding joint variable $q_j$, and compute the parent-frame transforms $\{T_w^{j-1}\}_{j=0}^{J}$ once via forward kinematics. Each occupied voxel of the retrieved kernels is then transformed to the world frame by its $T_w^{j-1}$ and tested against $\mathcal{K}_{map}$ via a single bit lookup, returning collision on the first hit. With bit-packed storage and early exit, each query runs in $O(|\mathcal{K}_{robot}|)$ in the worst case but terminates much sooner in practice, enabling high-throughput collision checking in the front-end search.

\section{Mid-end Trajectory Preprocessing} \label{sec:mid-end}
Directly solving the full whole-body optimization with the SVSDF safety term is expensive. We therefore first run a lightweight mid-end optimization that defers this term to the back-end: trusting the front-end path for collision safety, it enforces only smoothness and kinematic feasibility. Taking the discrete front-end path as input, it produces a continuous trajectory that is executed directly when collision-free, or otherwise passed to the back-end as a warm initial guess.
\subsection{Optimization Problem Formulation}
Following the constrained optimal control formulation, we cast the mid-end problem as:
\begin{subequations}\label{eq:midend_problem}
\begin{align}
&\min_{\{\mathbf{c}_{i,j}\},\, \{\mathbf{T}_i\}} \ J_0 = \int_{0}^{T_{\text{total}}} \!\! \psi(t)^\top W \psi(t)\, dt + \rho_T T_{\text{total}}, \label{eq:midend_obj} \\
&\text{s.t.}\quad \psi(t) = z^{(s)}(t),  \forall t \in [0, T_{\text{total}}], \label{eq:midend_psi} \\
&z^{[s-1]}(0) = \bar{z}_0,  z^{[s-1]}(T_{\text{total}}) = \bar{z}_f, \label{eq:midend_boundary} \\
&z_i^{[s-1]}(T_i) = z_{i+1}^{[s-1]}(0),  i = 1, \ldots, N-1, \label{eq:midend_continuity} \\
&\mathcal{C}_{d_b}\!\left(z(t), \ldots, z^{(s)}(t)\right) \leq 0, \forall d_b \in \mathcal{D}_b, \ \forall t \in [0, T_{\text{total}}], \label{eq:midend_base} \\
&\mathcal{C}_{d_m}\!\left(z(t), \ldots, z^{(s)}(t)\right) \leq 0, \forall d_m \in \mathcal{D}_m, \ \forall t \in [0, T_{\text{total}}], \label{eq:midend_joint}
\end{align}
\end{subequations}
where $W \in \mathbb{R}^{(2+J) \times (2+J)}$ is a diagonal matrix penalizing control effort and $\rho_T T_{\text{total}}$ is the time regularization term. Equation~\eqref{eq:midend_psi} defines the control input $\psi(t)$ as the $s$-th derivative of the trajectory, which serves as the integrand of the control-effort cost in~\eqref{eq:midend_obj}. Equations~\eqref{eq:midend_boundary} and~\eqref{eq:midend_continuity} impose the boundary conditions and the $C^{s-1}$ continuity at segment connections, respectively, where $\bar{z}_0$ and $\bar{z}_f$ denote the prescribed initial and terminal states. The remaining inequality constraints~\eqref{eq:midend_base} and~\eqref{eq:midend_joint} encode the differential base constraints $\mathcal{D}_b = \{d_b : \omega_{l(r)}, \alpha_{l(r)}, e_b\}$ and manipulator constraints $\mathcal{D}_m = \{d_m : q_j, \dot{q}_j, \ddot{q}_j, e_m\}$, where $e_b$ and $e_m$ are the pose residuals to the front-end path explained in detail in Sec.~\ref{sec:const}. In MINCO~\cite{wang2022minco} trajectories, the piecewise polynomial coefficients $\{\mathbf{c}_{i,j}\}$ are uniquely determined by a set of intermediate waypoint parameters $\mathbf{Q}\in\mathbb{R}^{(2+J)\times(\sum_i M_i-1)}$ together with the time allocations $\{\mathbf{T}_i\}$. We therefore reparameterize the optimization variables from $\{\mathbf{c}_{i,j}\}$ to $\mathbf{Q}$, under which the boundary conditions~\eqref{eq:midend_boundary} and the $C^{s-1}$ continuity constraints~\eqref{eq:midend_continuity} are satisfied by construction.

\subsection{Mid-end Constraints} \label{sec:const}
 The inequality constraints in~\eqref{eq:midend_base} and~\eqref{eq:midend_joint} are reformulated into penalty functions by constraint transcription method~\cite{constraint_elimination}. To prevent the mid-end trajectory from deviating excessively from the front-end path, we introduce pose residual constraints that enforce proximity to the discrete full-state path provided by the front-end, where each node is given by $x = [x_b^\top, x_m^\top]^\top$, $x_b$ contains the base position $p_x,p_y$ and orientation $\phi_b$, and $x_m$ contains the manipulator joint positions $q_j$. The residual constraints are defined as:
\begin{subequations}\label{eq:pose_residual}
\begin{align}
\mathcal{C}_{e_b}(z(t)) &= \|p_b(t) - p_{b,k(t)}\|^{2} + w_\phi\, \angle\!\big(\phi_b(t), \phi_{b,k(t)}\big)^{2} - \rho_b^{2}, \\
\mathcal{C}_{e_m}(z(t)) &= \|x_m(t) - x_{m,k(t)}\|^{2} - \rho_m^{2},
\end{align}
\end{subequations}
where $\|\cdot\|^2$ denotes the squared Euclidean norm of a vector, $k(t)$ indexes the nearest front-end node to time $t$, $\angle(\cdot,\cdot)$ denotes the shortest signed angular difference on $\mathbb{S}^1$, $w_\phi > 0$ is a scalar weight balancing the angular and translational contributions, and $\rho_b$, $\rho_m$ are the admissible deviation bounds for the base pose and manipulator joint configuration, respectively. Additionally, to ensure that the trajectory does not exceed the tracking capabilities of the mobile manipulator, we incorporate kinematic constraints on both the mobile base and the manipulator. For the mobile base, we restrict the angular velocity $\omega_{l(r)}$ and acceleration $\alpha_{l(r)}$ of the left and right wheels to their maximum admissible values. For each manipulator joint $j \in \{1, \ldots, J\}$, we impose limits on joint position $q_j$, velocity $\dot{q}_j$, and acceleration $\ddot{q}_j$.


\section{BACKEND TRAJECTORY OPTIMIZATION} \label{sec:back-end}
Whereas the mid-end in Sec.~\ref{sec:mid-end} trusts the front-end path and only penalizes deviation from it, the back-end solves the complete optimization problem in Eq.~\eqref{eq:midend_problem}, explicitly enforcing continuous collision avoidance for the mobile manipulator together with its grasped payload of arbitrary shape. To this end, we first review the swept volume signed distance field (SVSDF)~\cite{SVSDF} in Sec.~\ref{sec:svsdf} and identify its limitation when applied directly to a kinematically coupled multi-link system. We then propose the \emph{Kinematically-Coupled SVSDF} (KC-SVSDF) in Sec.~\ref{sec:kc-svsdf} to address this limitation, and finally formulate the back-end safety penalty in Sec.~\ref{sec:safety_constraints}.

\subsection{Swept Volume SDF}\label{sec:svsdf}
For continuous, mesh-level collision evaluation along an optimized trajectory, we build on the swept volume signed distance field (SVSDF) proposed in~\cite{SVSDF}. Given a rigid body with geometry $\mathcal{R}$ moving along a trajectory $\mathcal{T}(t)$, the swept volume is the union of all configurations the body occupies along the motion:
\begin{equation}\label{eq:sv}
\mathcal{SV} \triangleq \bigcup_{t \in [t_{\text{start}},\, t_{\text{end}}]} \mathcal{T}(t)\,\mathcal{R},
\end{equation}
where $\mathcal{T}(t)$ is a homogeneous rigid transformation, so $\mathcal{T}(t)\,\mathcal{R}$ denotes the body configuration at time $t$. The SVSDF at a query point $\bm{p}$ is the signed distance from $\bm{p}$ to the boundary $\mathcal{FR}(\mathcal{SV})$; equivalently, $|\mathcal{SVSDF}(\bm{p})|$ is the radius of the largest open ball $\mathcal{B}_{\bm{p}}(r)$ centered at $\bm{p}$ that does not intersect $\mathcal{FR}(\mathcal{SV})$, with the sign taken positive outside $\mathcal{SV}$ and negative inside.

Computing this radius directly is difficult because $\mathcal{SV}$ is an infinite union with no closed-form boundary. However, since $\mathcal{SV}$ is the union over time of instantaneous configurations $\mathcal{T}(t)\,\mathcal{R}$, the distance from $\bm{p}$ to $\mathcal{SV}$ equals the minimum, over $t$, of its distances to each individual configuration. Each such per-instant distance is in turn obtained by transforming $\bm{p}$ into the body's local frame via $\mathcal{T}^{-1}(t)\,\bm{p}$ and querying the static SDF on $\mathcal{R}$. This reduces the ball-radius problem to a one-dimensional minimization over time, and following~\cite{SVSDF} we adopt the resulting metric function
\begin{equation}\label{eq:g}
g(\bm{p}) \triangleq \min_{t \in [t_{\text{start}},\, t_{\text{end}}]} \mathcal{SDF}^{\mathcal{R}}\!\left(\mathcal{T}^{-1}(t)\,\bm{p}\right),
\end{equation}
which is a conservative SDF of $\mathcal{SV}$: $g(\bm{p}) = \mathcal{SVSDF}(\bm{p})$ when $\bm{p}$ lies outside $\mathcal{SV}$, and $g(\bm{p}) \geq \mathcal{SVSDF}(\bm{p})$ when $\bm{p}$ lies inside. The associated argmin time is denoted as
\begin{equation}\label{eq:tstar}
t^{*}(\bm{p}) \triangleq \operatorname*{arg\,min}_{t \in [t_{\text{start}},\, t_{\text{end}}]} \mathcal{SDF}^{\mathcal{R}}\!\left(\mathcal{T}^{-1}(t)\,\bm{p}\right).
\end{equation}

The SVSDF is formulated for a single rigid body. A mobile manipulator carrying a grasped object, however, is a kinematic chain $\mathcal{M} = \{\mathcal{R}_{0}, \mathcal{R}_{1}, \ldots, \mathcal{R}_{J}, \mathcal{R}_{J+1}\}$ comprising the mobile base ($\mathcal{R}_{0}$), the manipulator links ($\mathcal{R}_{1}, \ldots, \mathcal{R}_{J}$), and the grasped payload ($\mathcal{R}_{J+1}$) rigidly attached to the end-effector. A natural extension would be to construct a separate swept volume for each rigid body $\mathcal{R} \in \mathcal{M}$ over the trajectory segment $\mathcal{T}(t_i, t_{i+1})$ and compute its SVSDF independently. This per-body treatment, however, implicitly assumes that each body can move on its own. For a serial manipulator this assumption fails: realizing link $j$'s obstacle escape direction requires actuating upstream joints that simultaneously displace other rigidly connected links, which could produce mutually contradictory per-link gradients that stall the optimizer. We resolve this by coupling the per-link distance fields through the kinematic chain, as introduced in Sec.~\ref{sec:kc-svsdf}.

\subsection{Kinematically-Coupled SVSDF}\label{sec:kc-svsdf}


\begin{figure*}[t]
    \centering
    \includegraphics[width=0.98\textwidth]{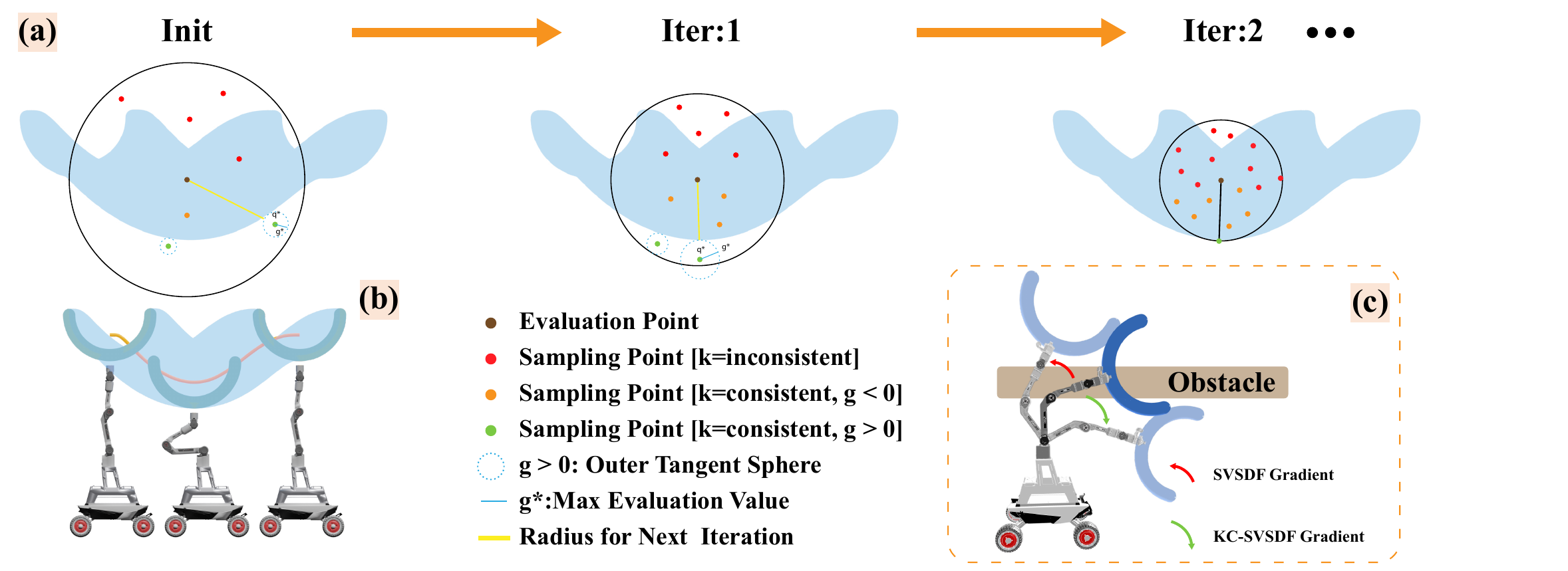}
    \captionsetup{font={footnotesize}}
    \caption{(a) Iterative computation of the interior KC-SVSDF for the payload link, illustrated in 2D (the 3D case is analogous). Candidate samples on the current ball are filtered by \textsc{KinematicFilter} for upstream-joint consistency with $\bm{G}$; surviving samples (green) outside $\mathcal{SV}$ yield metric values $g$ (tangent-circle radii). The worst violator $g^{*}$ shrinks the ball radius for the next iteration, rapidly converging to the KC-SVSDF value inside $\mathcal{SV}$. (b) Swept volume (light blue) generated by the mobile manipulator transporting an arc-shaped payload along a trajectory. (c) When the payload collides with an obstacle, the traditional SVSDF produces a per-link escape direction (red) that ignores the conflicting, whereas our KC-SVSDF propagates the gradient through the chain to yield a kinematically consistent whole-body escape direction (green).}
    \label{fig:kc-svsdf-iter}
    \vspace{-10pt}
\end{figure*}

To resolve the per-link inconsistency identified in Sec.~\ref{sec:svsdf}, we propose the KC-SVSDF, summarized in Algorithm~\ref{alg:kc-svsdf} and illustrated in Fig.~\ref{fig:kc-svsdf-iter}. The procedure walks down the kinematic chain link by link, computing each link's SVSDF while constraining its escape direction to be consistent with the directions already chosen by upstream links. Consistency is tracked through a joint-space accumulator $\bm{G} \in \mathbb{R}^{n_q}$, reset to zero at the start of each query and updated after each link's SVSDF is determined.
\begin{algorithm}
\caption{KC-$\mathcal{SVSDF}$ Computation}\label{alg:kc-svsdf}
\begin{algorithmic}[1]
\State \textbf{Input:} \textit{query point} $\bm{p}$
\State \textbf{Output:} \textit{per-body KC-SVSDF values} $\bm{D} \in \mathbb{R}^{J+2}$
\State $\bm{G} \gets \bm{0} \in \mathbb{R}^{n_q}$
\State $\bm{D} \gets \bm{0} \in \mathbb{R}^{J+2}$
\For{$j = 0$ to $J+1$}
    \State $g_j(\bm{p}) \gets \Call{SVSDF}{\mathcal{R}_{j}, \bm{p}, \mathcal{T}}$
    \If{$g_j(\bm{p}) > 0$}
        \State $\nabla_q^{(j)} \gets -\mathbf{J}_j^{\top}\,\nabla g_j(\bm{p})$
        \State $\bm{G}[0{:}j] \gets \bm{G}[0{:}j] + \nabla_q^{(j)}[0{:}j]$
        \State $\bm{D}[j] \gets g_j(\bm{p})$
        \State \textbf{continue}
    \EndIf
    \State $r \gets$ \textit{a big initial value}
    \Loop
        \State $Y \gets$ uniform samples on the surface of $\mathcal{B}_{\bm{p}}(r)$
        \State $\bm{y} \gets \Call{KinematicFilter}{Y, \bm{G}, j}$
        \State $g_j(\bm{y}^*) \gets \arg\max g_j(\bm{y})$
        \If{$g_j(\bm{y}^*) < \epsilon$}
            \State $\nabla_q^{(j)} \gets \mathbf{J}_j^{\top}(\bm{y}^* - \bm{p})$
            \State $\bm{G}[0{:}j] \gets \bm{G}[0{:}j] + \nabla_q^{(j)}[0{:}j]$
            \State $\bm{D}[j] \gets -r$
            \State \textbf{break}
        \EndIf
        \State $r \gets r - g_j(\bm{y}^*)$
    \EndLoop
\EndFor
\State \Return $\bm{D}$
\end{algorithmic}
\end{algorithm}

For each link $j$, we first evaluate $g_j(\bm{p})$ as defined in Sec.~\ref{sec:svsdf} (line~6). If $\bm{p}$ lies outside the swept volume ($g_j(\bm{p}) > 0$), the KC-SVSDF is taken as $g_j(\bm{p})$, and the spatial escape direction $-\nabla g_j(\bm{p})$ is mapped to joint space through link $j$'s geometric Jacobian $\mathbf{J}_j$ and accumulated into $\bm{G}[0{:}j]$ (lines~8--10). Otherwise, we run the iterative ball-shrinking loop shown in Fig.~\ref{fig:kc-svsdf-iter}: candidate samples on the ball surface (line~15) are filtered by the kinematic-consistency rule (line~16)
\begin{equation}\label{eq:kc-svsdf-filter}
\nabla_q^{(k)}[i] \cdot \bm{G}[i] \geq 0, \quad \forall i \in \{j{-}2,\, j{-}1\},
\end{equation}
which early rejects any candidate whose induced motion would push the two immediately upstream joints against the consensus already established in $\bm{G}$ (the check is skipped for $j < 2$) and do not evaluate its $g_j$ to improve the efficiency. Among the surviving samples $\bm{y}$, the worst violator $g_j$ in sample $\bm{y}^*$ drives the radius update $r \gets r - g_j$ (line~24), and the loop terminates when $g^{*} < \epsilon$, at which point $-r$ is recorded as link $j$'s KC-SVSDF (lines~21).


The resulting KC-SVSDF value $\bm{D}$ encodes coherent whole-body escape directions that the kinematic chain can physically realize, as visualized in Fig.~\ref{fig:kc-svsdf-iter}(c): where independent per-link SVSDF gradients (red) would drive upstream links back into collision, the KC-SVSDF gradients (green) yield a consistent whole-arm motion that clears the obstacle.



Computing the KC-SVSDF for mobile manipulator requires solving the argmin-over-time search of Eq.~\eqref{eq:tstar} once per link, which is costly across the $J{+}2$ links and the many query points evaluated each iteration. We adopt a strategy that exploits the rigid coupling of the chain to avoid this. Since the manipulator links and the grasped payload are carried by the base along a shared trajectory, the base's translation dominates their spatial motion, so each link's closest-approach time $t^{*}(\bm{p})$ falls within a small, bounded offset of the base's argmin time $t_{0}^{*}(\bm{p})$. We therefore solve Eq.~\eqref{eq:tstar} in full only for the base, and confine the search for each remaining link $\{\mathcal{R}_{j}\}_{j=1}^{J+1}$ to the narrow window $[t_{0}^{*}(\bm{p}) - \Delta t,\ t_{0}^{*}(\bm{p}) + \Delta t]$, whose half-width $\Delta t$ covers this offset. This turns $J{+}2$ full-segment searches into one full search plus $J{+}1$ windowed searches, substantially reducing the per-query cost.

\subsection{Safety Constraints} \label{sec:safety_constraints}

The back-end shares the same trajectory parameterization, smoothness objective, and kinematic feasibility constraints as the mid-end in Eq.~\eqref{eq:midend_problem}, and additionally enforces continuous collision avoidance for the whole-body system through a KC-SVSDF-based safety penalty. The goal is to ensure that the swept volume $\mathcal{SV}$ of every link in mobile manipulator remains clear of obstacles along the entire trajectory. Building on the per-link KC-SVSDF values $\bm{d}$ stores in $\bm{D}$ computed in Sec.~\ref{sec:kc-svsdf}, we require that, for every obstacle point near the trajectory, the $d$ exceeds a prescribed safety margin $s_{\text{thr}}$. The resulting penalty is
\begin{subequations}\label{eq:safety_penalty}
\begin{align}
J_s &= \sum_{i=1}^{N_{\text{obs}}} \sum_{j=0}^{J+1} \mathcal{L}\!\left(\mathcal{G}_s\!\left(\bm{d}_j(\bm{x}_{\text{obs}}^{\,i})\right)\right), \\
\mathcal{G}_s(d) &= \max\!\left(\,0,\ s_{\text{thr}} - d\,\right),
\end{align}
\end{subequations}
where $\bm{d}_j(\bm{x}_{\text{obs}}^{\,i})$ denotes the KC-SVSDF of link $j$ at obstacle point $\bm{x}_{\text{obs}}^{\,i}$, $\mathcal{L}(\cdot)$ is a smooth penalty function (e.g., a cubic barrier) that converts the violation into a differentiable cost, and $s_{\text{thr}} > 0$ is the user-specified safety margin. The candidate obstacle points $\{\bm{x}_{\text{obs}}^{\,i}\}_{i=1}^{N_{\text{obs}}}$ are pre-filtered by an Axis-Aligned Bounding Box (AABB) test around the trajectory, so that only obstacle voxels potentially intersecting any link's swept volume are evaluated, while the kinematically-coupled gradients propagated through $\bm{d}$ ensure that the back-end produces a continuously collision-free trajectory for the mobile manipulator and its arbitrarily shaped payload.

\section{Experiments}
We validate the proposed framework through both simulation and real-world experiments. The simulation studies investigate the contribution of KC-SVSDF module via an ablation study and benchmark the full system against state-of-the-art baselines, while the real-world experiments demonstrate the framework's deployability and robustness on a physical platform. All simulations and real-world experiments are performed in the Robot Operating System (ROS) on an Intel NUC with an i7-1360P CPU and 32\,GB DDR4 memory.

\begin{table}[h]
\vspace{-10pt}
\centering
\caption{Ablation Success Rate Results}
\label{tab:ablation}
\begin{tabular}{ccccc}
\toprule
$h_b$ (m) & 1.1 & 1.0 & 0.9 & 0.8 \\
\midrule
Baseline & 80\% & 35\% & 20\% & 0.00\% \\
Ours     & \textbf{100\%} & \textbf{100\%} & \textbf{95\%} & \textbf{80\%} \\
\bottomrule
\end{tabular}
\vspace{-10pt}
\end{table}

\subsection{Simulation Experiments}

\subsubsection{Comparative Benchmark}
\begin{figure*}[t]
    \centering
    \includegraphics[width=0.98\textwidth]{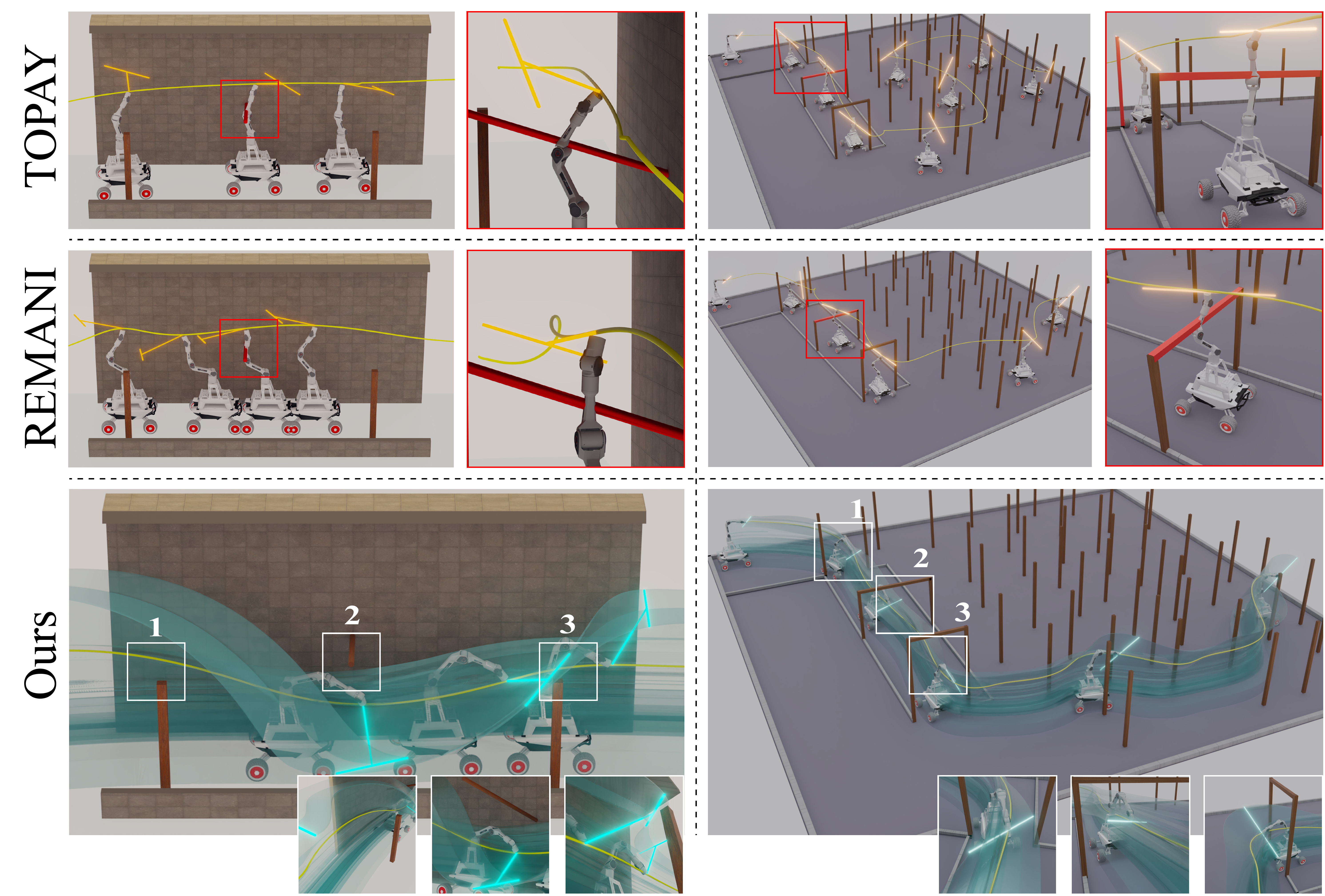}
    \captionsetup{font={footnotesize}}
    \caption{Qualitative comparison in two challenging environments: a narrow tunnel (left) and a dense forest of vertical rods (right). The number above each white box in bottom row indicates the corresponding obstacle in Table~\ref{tab:simulation_kcsvsdf}. The robot is a differential-drive mobile manipulator with a 6-DOF arm carrying a non-convex T-shaped and an elongated rectangular payload. Red boxes (TOPAY, REMANI) highlight payload-obstacle collisions. Our method (bottom row) plans over the kinematically-coupled swept volume of the full robot-payload system (teal envelope), achieving collision-free navigation through tight clearances in both scenarios.}
    \label{fig:simulation_benchmark}
    \vspace{-10pt}
\end{figure*}

We compare our full method against two state-of-the-art trajectory-optimization baselines for differential-drive mobile manipulators, REMANI~\cite{remani} and TOPAY~\cite{topay}, across two scenarios of increasing difficulty: \textbf{Tunnel}, a structured corridor with regularly placed obstacles (Fig.~\ref{fig:simulation_benchmark}, left); and \textbf{Forest}, a cluttered environment containing three bridge holes and forty-five randomly placed columns (Fig.~\ref{fig:simulation_benchmark}, right). The Tunnel scenario uses the same T-shaped payload as the ablation study, while the Forest scenario uses an elongated rectangular payload to stress-test the planner with a larger grasped object. 

\begin{table}[h]
\centering
\caption{Minimum SVSDF at Challenging Obstacles}
\label{tab:simulation_kcsvsdf}
\setlength{\tabcolsep}{5pt}
\begin{tabular}{llccc}
\toprule
 & Method & Obstacle 1 & Obstacle 2 & Obstacle 3 \\
\midrule
\multirow{3}{*}{Tunnel}
 & Ours   & \textbf{0.3145} & \textbf{0.1882} & \textbf{0.0714} \\
 & TOPAY  & 0.2234          & -0.0080          & 0.3438          \\
 & REMANI & 0.1477          & -0.0390          & 0.3875          \\
\midrule
\multirow{3}{*}{Forest}
 & Ours   & \textbf{0.0706} & \textbf{0.1069} & \textbf{0.1598} \\
 & TOPAY  & 0.0573          & -0.0317          & 0.1167          \\
 & REMANI & 0.1764          & -0.0077          & 0.1855          \\
\bottomrule
\end{tabular}
\vspace{-10pt}
\end{table}

For each scenario, we report over 100 trials the success rate and the average per-module computation time in Table~\ref{tab:simulation_benchmark}, together with the minimum SVSDF value at the most three challenging obstacles in Table~\ref{tab:simulation_kcsvsdf}, which are labeled in Fig.~\ref{fig:simulation_benchmark}. The minimum SVSDF denotes the closest distance between any robot link mesh and these obstacles across the entire trajectory. A trial is counted as a success when the robot reaches its global destination without collision. A positive SVSDF value indicates that every robot link stays outside the obstacles, whereas a negative value reports the penetration depth.
Our method achieves the highest success rate and the lowest total planning time in both scenarios. This performance gap stems from two compounding factors. First, both TOPAY and REMANI rely on sphere-based collision checking, which overapproximates the payload geometry and needlessly rejects feasible configurations in the front-end; this yields more planning attempts and poorer initial values for the back-end optimizer. Second, as illustrated in Fig.~\ref{fig:simulation_benchmark}, neither baseline accounts for the rigid coupling between manipulator links, so when the robot carries a large payload, freeing one link drives an upstream link into collision during trajectory optimization, ultimately producing low success rates.

\subsubsection{Ablation Study}
We conduct an ablation study to validate the KC-SVSDF safety formulation proposed in Sec.~\ref{sec:kc-svsdf}. The baseline replaces the KC-SVSDF safety term in the back-end with an independent per-link SVSDF~\cite{SVSDF}, while keeping the front-end, mid-end, and all other components of the pipeline identical to the full method. This isolates the contribution of the kinematic coupling and rules out improvements arising from other modules.

\begin{figure}[h]
    \centering
    \includegraphics[width=0.48\textwidth]{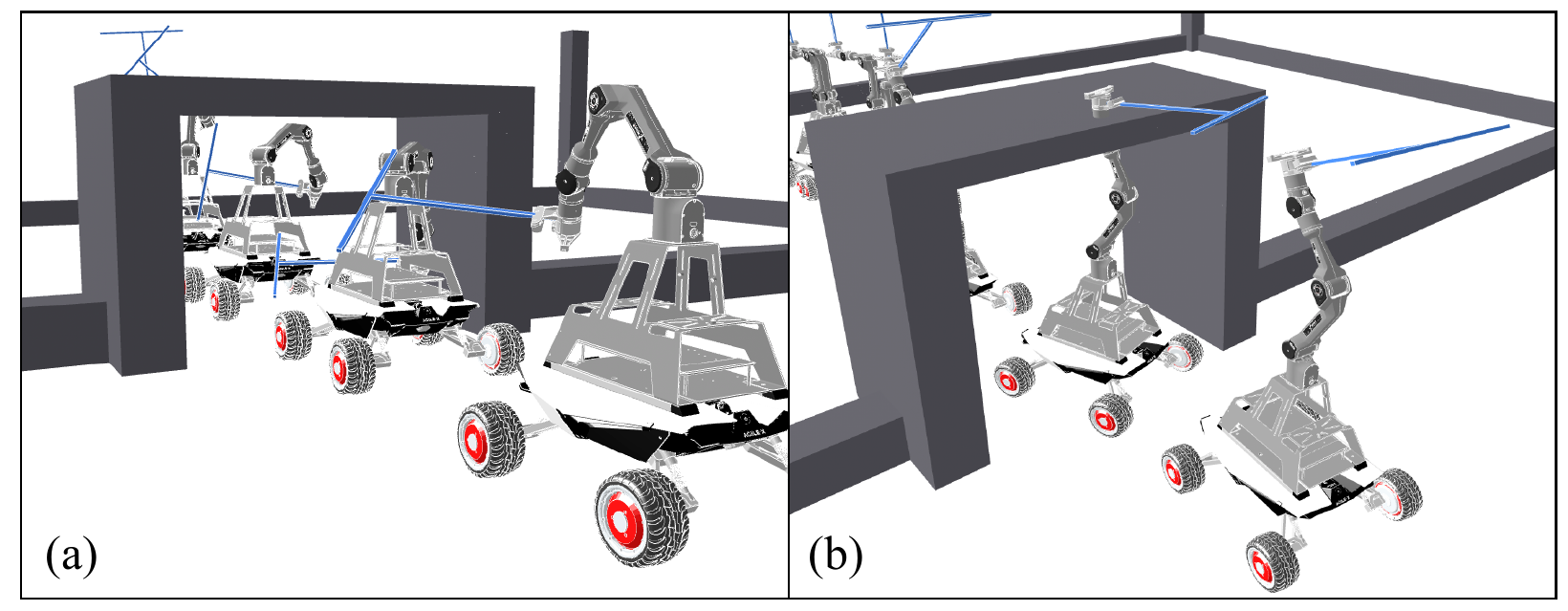}
    \captionsetup{font={footnotesize}}
    \caption{The generated trajectory in ablation study. (a): With KC-SVSDF $h_b$ = 0.8m. (b):SVSDF-only $h_b$ = 1.0m.}
    \label{fig:ablation}
    \vspace{-10pt}
\end{figure}

The scenario is a bridge-hole passage with a T-shaped payload rigidly grasped by the end-effector (Fig.~\ref{fig:ablation}). To probe the boundary of each method's capability, we sweep the bridge clearance height $h_b$ from $1.1$\,m down to $0.8$\,m in steps of $0.1$\,m. For each setting, we run 20 randomized trials and report the success rate, defined as completing the traversal without collision and within the time budget. The front-end path searching is limited to 80ms and the optimization module limited to 300ms. Results are summarized in Table~\ref{tab:ablation}.

\begin{table}[h]
\centering
\caption{Simulation Benchmark}
\label{tab:simulation_benchmark}
\setlength{\tabcolsep}{3pt}
\begin{tabular}{llccccc}
\toprule
 & Method & \shortstack{Front-end\\(ms)} & \shortstack{Mid-end\\(ms)} & \shortstack{Back-end\\(ms)} & \shortstack{Total\\(ms)} & \shortstack{Succ\\rate} \\
\midrule
\multirow{3}{*}{Tunnel}
 & Ours   & \textbf{2305} & \textbf{604}  & \textbf{1253}  & \textbf{4162}   & \textbf{91\%} \\
 & TOPAY  & 3345          & $-$           & 4533           & 7878            & 23\% \\
 & REMANI & 9186          & $-$           & 2039           & 11225           & 29\% \\
\midrule
\multirow{3}{*}{Forest}
 & Ours   & \textbf{9024} & \textbf{1022} & \textbf{3567}  & \textbf{13613}  & \textbf{83\%} \\
 & TOPAY  & 8771          & $-$           & 14407        & 23178         & 25\% \\
 & REMANI & 24233         & $-$           & 7207           & 31440           & 17\% \\
\bottomrule
\end{tabular}
\vspace{-20pt}
\end{table}

\subsection{Real-World Experiments}
Our hardware platform consists of an Agilex Piper 6-DOF manipulator mounted on an Agilex Scout Mini differential-drive base. State estimation for the base is provided by a Livox MID-360 LiDAR running FAST-LIO2~\cite{fastlio2}, while the manipulator joint states are read directly from the motor encoders. The 3D occupancy grid is maintained and updated online using ROG-Map~\cite{rogmap}.

\begin{figure}[t]
    \centering
    \includegraphics[width=0.48\textwidth]{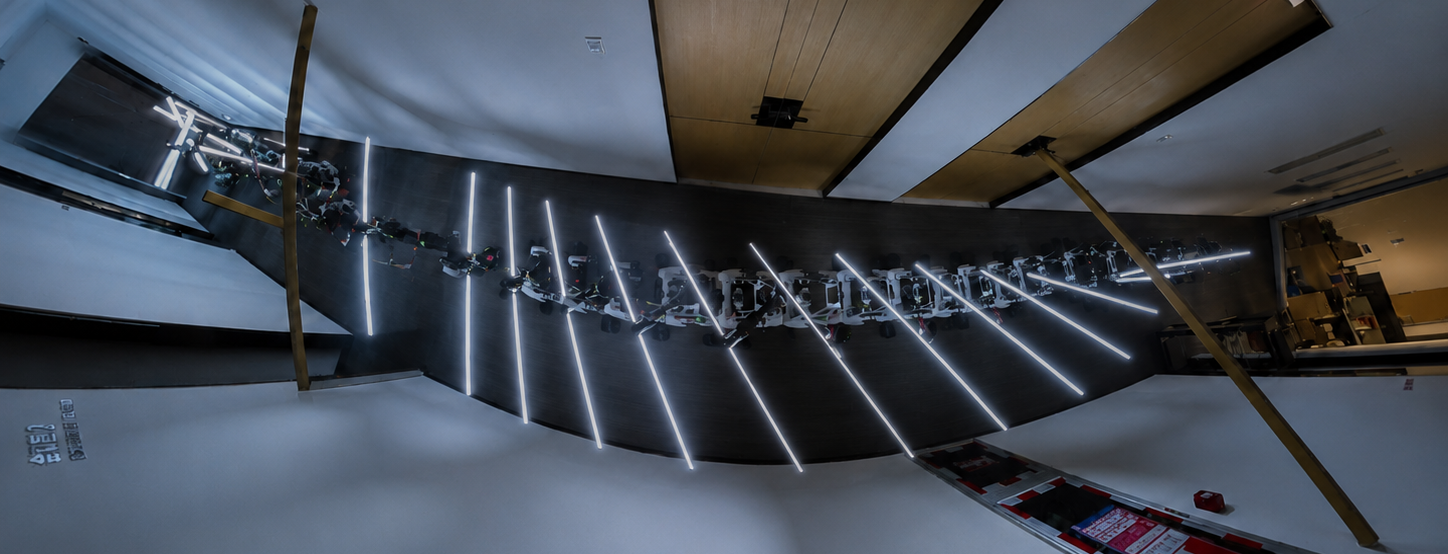}
    \captionsetup{font={footnotesize}}
    \caption{The mobile manipulator transports an elongated rectangular payload through a narrow corridor containing two inclined beam obstacles at different orientations and one vertical obstacle.}
    \label{fig:exp2}
    \vspace{-10pt}
\end{figure}

\begin{table}[h]
\centering
\caption{Results of the Real-world Experiment}
\label{tab:real_world_experiment}
\setlength{\tabcolsep}{5pt}
\begin{tabular}{lccccc}
\toprule
Scene
 & \shortstack{Front-end\\(ms)}
 & \shortstack{Mid-end\\(ms)}
 & \shortstack{Back-end\\(ms)}
 & \shortstack{Total\\(ms)}
 & \shortstack{Length\\(meter)} \\
\midrule
Tunnel & \textbf{523.6} & \textbf{3464.3}  & \textbf{5388}  & \textbf{9375.9}  & \textbf{19.477} \\
Forest & \textbf{782.6} & \textbf{4312.8} & \textbf{6346.2}  & \textbf{11441.6} & \textbf{17.256} \\
\bottomrule
\end{tabular}
\vspace{-10pt}
\end{table}

We build two complex scenarios on the physical platform to validate the framework on real hardware platform. In the first experiment shown in Fig.~\ref{fig:exp2}, the manipulator grasps an elongated rectangular payload and traverses a narrow tunnel, demonstrating safe whole-body navigation with a large object under tight geometric constraints. In the second experiment shown in Fig~\ref{fig:cover_page}, the manipulator carries a T-shaped payload through a forest of randomly placed wooden columns, demonstrating real-time re-planning in cluttered, less structured environments, the final result is listed in Table~\ref{tab:real_world_experiment}.


\section{CONCLUSIONS}

This letter presented a real-time whole-body planner for mobile manipulators transporting arbitrarily shaped payloads. Preserving true geometry recovers feasible space lost to sphere-based approximations, while Kinematically-Coupled SVSDF resolves the gradient inconsistency stalling per-link swept-volume optimization. The framework outperforms state-of-the-art baselines in success rate and computation time, and reliably transports large, non-convex payloads in real-world tests. Like all optimization-based planners, it remains sensitive to local minima—especially for multi-link rigidly coupled systems—depending on the front-end initial guess; future work will incorporate topological path search to provide diverse initial guesses toward the global optimum.







\bibliographystyle{IEEEtran}
\bibliography{references}

\end{document}